\documentclass[sigconf]{acmart}
\usepackage{multirow}
\usepackage{enumitem}
\usepackage{algorithm}
\usepackage{algorithmic}
\usepackage{booktabs}
\usepackage{siunitx}
\AtBeginDocument{%
  }

\copyrightyear{2026}
\acmYear{2026}
\setcopyright{cc}
\setcctype{by}
\acmConference[RecSys '26]{20th ACM Conference on Recommender Systems}{September 27-October 02, 2026}{Minneapolis, MN, USA}
\acmBooktitle{20th ACM Conference on Recommender Systems (RecSys '26), September 27-October 02, 2026, Minneapolis, MN, USA}
\acmDOI{10.1145/3773078.3831798}
\acmISBN{979-8-4007-2284-4/2026/09}

\begin{document}

\title{Residual Dominance as a Structural Account of\\ Last-Item Reliance in Causal Self-Attention Recommenders}

\author{Keito Kozaki}
\affiliation{%
  \institution{Hokkaido University}
  \city{Sapporo}
  \state{Hokkaido}
  \country{Japan}}
  \email{kozaki@lmd.ist.hokudai.ac.jp}

\author{Keigo Sakurai}
\affiliation{%
  \institution{Hokkaido University}
  \city{Sapporo}
  \state{Hokkaido}
  \country{Japan}}
\email{sakurai@lmd.ist.hokudai.ac.jp}

\author{Ren Togo}
\affiliation{%
  \institution{Hokkaido University}
  \city{Sapporo}
  \state{Hokkaido}
  \country{Japan}}
  \email{togo@lmd.ist.hokudai.ac.jp}

\author{Takahiro Ogawa}
\affiliation{%
 \institution{Hokkaido University}
 \city{Sapporo}
  \state{Hokkaido}
  \country{Japan}}
 \email{ogawa@lmd.ist.hokudai.ac.jp}

\author{Miki Haseyama}
\affiliation{%
  \institution{Hokkaido University}
  \city{Sapporo}
  \state{Hokkaido}
  \country{Japan}}
  \email{mhaseyama@lmd.ist.hokudai.ac.jp}




\renewcommand{\shortauthors}{Kozaki et al.}

\begin{abstract}

Transformer-based sequential recommenders with causal self-attention often rely heavily on the most recent interaction at inference time, but how this behavior is structurally expressed in the representation
used for prediction remains unclear. We combine prediction-time diagnostics with norm-based analysis of the full attention block. First, we show that SASRec-style models exhibit highly localized last-item reliance. We then find that, although self-attention aggregates contextual information, residual addition sharply shifts the full-block representation toward same-position contributions, which we term residual dominance. To probe this interpretation, we use inference-time residual scaling as a controlled diagnostic intervention. Changing the residual strength induces a monotonic trade-off between structural mixing and last-item reliance, while reducing residual strength recovers a subset of final-position misses for which representations at non-final positions already rank the ground-truth item correctly. Our results provide a structural account linking extreme
last-item reliance to residual dominance at inference time.
The code is publicly available.\footnote{\url{https://github.com/keito0329/Residual}}
\end{abstract}




\maketitle

\section{Introduction}

\begin{figure*}[h]
  \centering
  \includegraphics[width=\linewidth]{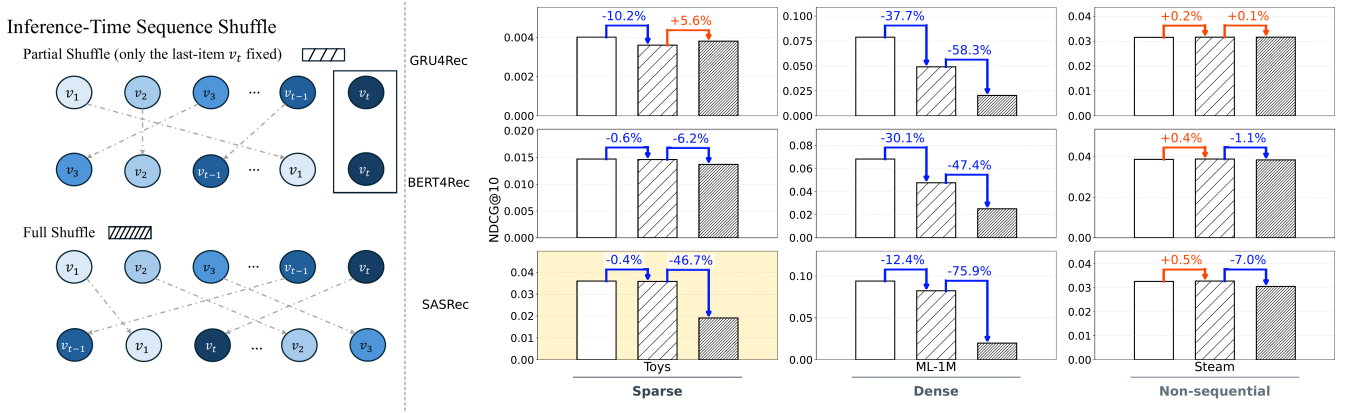}
  \caption{Impact of sequence ordering at inference time. Empty bars: chronologically ordered sequences; light diagonal hatch patterns: \emph{partial shuffle} where only the final position $v_t$ is kept fixed and the remaining items are permuted; heavy diagonal hatch patterns: \emph{full shuffle} where all positions, including $v_t$, are permuted. \textbf{Key takeaway:} on the sparse and dense datasets, preserving only
$v_t$ causes a substantially smaller change in SASRec's performance
than full shuffling,
indicating a discontinuous last-position privilege; this contrast
is weak on the non-sequential dataset.}
\Description{
The figure consists of a sequence-shuffling diagram on the left and nine bar charts on the right. In the partial-shuffle diagram, the final item remains fixed at the final position while all preceding items are randomly permuted. In the full-shuffle diagram, all items, including the original final item, are permuted. The bar charts are arranged in three rows for GRU4Rec, BERT4Rec, and SASRec, and three columns for Toys, ML-1M, and Steam. Each chart compares NDCG at 10 for the original chronological sequence, partial shuffle, and full shuffle. For SASRec on Toys and ML-1M, partial shuffling produces a substantially smaller performance change than full shuffling. On Steam, both perturbations produce relatively small changes. GRU4Rec and BERT4Rec generally exhibit weaker contrasts between partial and full shuffling than SASRec.
}
  \label{fig:shuffle}
\end{figure*}

Sequential recommendation predicts~\cite{survey/fang2020deep, survey/wang2019sequential,survey/pan2026survey} a user’s next interaction from an ordered sequence of past interactions. Among many approaches~\cite{model/MC/rendle2010factorizing,model/MC/zimdars2013using,model/rnn/hidasi2015session,model/sun2019bert4rec}, Transformer~\cite{model/vaswani2017attention}-based models with causal self-attention~\cite{model/wu2020sse,model/BSARec/shin2024attentive}, exemplified by SASRec~\cite{model/kang2018self}, have become a dominant paradigm due to their strong empirical performance~\cite{analysis/klenitskiy2023turning,analysis/koneru2025sasrec}. These models typically generate recommendations from the final-position representation, implicitly assuming that it aggregates useful signals from the full interaction history. 
In this work, we revisit this assumption at inference time:
\emph{how is information from different positions actually expressed
in the representation used for prediction?}
Importantly, our goal is not to determine why a dataset or training
process makes the last item predictive, but to examine how a trained
causal self-attention recommender structurally expresses last-item
reliance at inference time.

Recent studies~\cite{analysis/xu2026markovian,analysis/oh2024measuring} suggest that causal self-attention recommenders often exhibit highly localized inference-time behavior, where predictions are strongly anchored to the most recent interaction. As illustrated in Figure~\ref{fig:shuffle}, preserving only the last item while perturbing earlier positions causes only minor changes, whereas also perturbing the final position leads to a sharp degradation. 
Complementary ranking diagnostics~\cite{analysis/oh2024measuring}
show that causal self-attention models frequently rank the last
interacted item first, while doing so far less often for immediately
preceding items.
Although these diagnostics capture different aspects of model behavior, together they indicate a discontinuous privilege of the final position.
While this behavior is consistent with the intuition that recent interactions are often informative, its extreme localization remains insufficiently explained.

If positional influence were determined mainly by attention weights, one might expect a gradual decay from the final position to nearby preceding positions~\cite{analysis/wang2025your}. Instead, we observe a discontinuous pattern: predictive influence collapses onto the final item, while immediately preceding positions contribute little. Prior explanations based on recency bias or attention patterns therefore remain largely descriptive and do not fully account for how learned signals are actually utilized at prediction time~\cite{analysis/wang2025your, analysis/xu2026markovian}. This suggests that components beyond attention weights play an important role in shaping inference behavior.

To study this question, we analyze the full attention block, including residual connections and normalization, using norm-based decomposition~\cite{analysis/kobayashi2020attention, analysis/kobayashi2021incorporating}. 
Our results show that, although the attention output aggregates
contextual information, adding the residual pathway sharply reduces the relative contribution of preceding positions and strongly preserves same-position information.
Because SASRec forms its prediction from the final-position
representation, this self-information-retention structure provides a
direct pathway through which the most recent item can dominate the
prediction representation.
We refer to this structural tendency as \emph{residual dominance}.
We do not interpret residual dominance as the sole origin of last-item reliance, but as an important inference-time mechanism through which learned last-item reliance is expressed.

To probe this interpretation, we employ inference-time residual scaling as a controlled diagnostic intervention.
Residual scaling provides a one-dimensional modification of the residual contribution while keeping the trained parameters fixed, allowing us to examine whether representation structure and prediction behavior vary systematically with residual strength without retraining.
This intervention is intended as a mechanistic sensitivity probe rather than strict causal identification or a new recommendation method.
Our results show that reducing residual strength weakens the direct
preservation of the final-position input, while recovering a subset
of previously missed predictions for which representations at non-final positions already rank the ground-truth item correctly under standard inference. These findings suggest that prediction errors can depend not only on what information is encoded, but also on how that information is expressed at inference time.

Overall, we provide a structural account of how residual dominance
is associated with last-item reliance and use residual scaling to
probe prediction-time information utilization.

In summary, the contributions of this work are as follows:
\begin{itemize}[leftmargin=1.1em]
    \item We systematically characterize last-item reliance in causal sequential recommenders using complementary prediction-time diagnostics across diverse datasets.

    \item We analyze the complete attention block and show that the residual pathway sharply reduces contextual mixing while preserving same-position information, providing a structural account of how the final item can dominate the representation used for prediction.

     \item We use inference-time residual scaling as a controlled diagnostic probe and show that residual strength is systematically associated with last-item reliance, contextual mixing, and the recovery of final-position misses with correct non-final predictions.
\end{itemize}

\section{Preliminaries and Overall Setup}
\label{sec:setup}
This section defines the task notation and summarizes the datasets, baselines, and evaluation protocol.
Unless otherwise specified, the settings described here are applied to all experiments and diagnostics in Sections~\ref{sec:empirical}--\ref{sec:Inferencetime}.

\subsection{Preliminaries}
Consider a user set $\mathcal{U} = \{u_1, u_2, \dots, u_{|\mathcal{U}|}\}$ and an item set $\mathcal{V} = \{v_1, v_2, \dots, v_{|\mathcal{V}|}\}$.
For a user $u$, we denote the chronological interaction sequence as $S_u=(v_1, v_2, \dots, v_t)$, where $t=|S_u|$ and $v_t$ is the most recent item.
The next-item prediction task is to estimate a probability distribution over $\mathcal{V}$ for the next item $v_{t+1}$:
\begin{equation}
    p(v_{t+1}=v \mid S_u; \Theta), \quad v\in\mathcal{V},
\end{equation}
where $\Theta$ represents the model parameters.
Throughout the remainder of the paper, we denote the final position by $L$
(i.e., $L = t$), and refer to preceding positions as $L-1$, $L-2$, etc.,
when discussing position-wise diagnostics.

\subsection{Overall Setup}

\begin{table}[t]
  \footnotesize
  \centering
  \caption{Statistics of the processed datasets after $p$-core filtering ($p=5$) and consecutive-repeat removal.}
  \label{tab:dataset_statistics}
  \scalebox{1}[1]{
    \begin{tabular}{lcccccc}
      \hline
      Dataset & \#Users & \#Items & \#Interact. & Avg. Len. & Density & \#Days\\
      \hline
      Beauty   & 22,363  & 12,101 & 198,502 & 8.9 & 0.07\% & 4,424\\
      Sports  & 35,598  & 18,357 & 296,337 & 8.3 & 0.05\% & 4,521\\
      Video & 24,303 & 10,672 & 231,780 & 9.5 &  0.09\% & 5,395\\
      Diginetica    & 61,279  & 25,593 & 485,903 & 7.9 & 0.03\% & 152\\
      Toys    & 19,412  & 11,924& 167,597 & 8.6& 0.07\% & 5,108\\
      Steam    & 281,349  & 11,961 & 3,550,272 & 12.6& 0.11\% & 2,639 \\
      BeerAdvocate    & 14,635  & 22,074 & 1,475,412 & 100.8& 0.46\% & 5,620\\
      ML-1M    & 6,040  & 3,416 & 999,611 & 165.5& 4.84\% & 1,038\\
      Zvuk & 19,267 & 150,206 & 8,087,953 & 419.8 & 0.28\% &91\\
      \hline
\end{tabular}
}
\end{table}

\subsubsection{Datasets}
\label{subsec:datasets}
We conduct our analysis on nine publicly available datasets that are widely used in sequential recommendation research and collectively cover a broad range of domains, sparsity levels, sequence lengths, and temporal characteristics. 

As summarized in Table~\ref{tab:dataset_statistics}, these datasets vary in scale and structure, spanning both short session-based interactions and long-term user histories. 
All interactions are treated as implicit feedback, following the standard next-item prediction setting~\cite{analysis/klenitskiy2023turning,model/caser/tang2018personalized}.

Importantly, the datasets used in this study are deliberately chosen to reflect different degrees of sequential structure discussed in recent literature~\cite{analysis/klenitskiy2024does}.  
Specifically, our benchmark includes datasets that are widely regarded as exhibiting clear sequential dependencies and being well suited for sequential recommendation, such as the Amazon Beauty, Sports, and Zvuk datasets, as well as datasets for which prior work~\cite{analysis/klenitskiy2024does} has reported weaker or more limited sequential structure, including Diginetica and Steam.
By covering datasets with substantially different degrees of inherent sequentiality, our experimental setting enables us to examine whether prediction-time information utilization, and in particular last-item reliance, consistently emerges across datasets, rather than being an artifact of a specific dataset choice.

To ensure data quality and comparability across datasets, we apply $5$-core filtering~\cite{analysis/sachdeva2020useful, analysis/ferrari2019we,analysis/sun2020we}.  
In addition, consecutive repeated items within user sequences are removed~\cite{analysis/hidasi2023widespread}, as such repetitions do not convey additional sequential dependency information and may artificially inflate recency effects.

\subsubsection{Baseline Models}
\label{subsec:baselines}

We consider three representative baseline models for sequential recommendation: GRU4Rec, SASRec, and BERT4Rec.
These models are widely used benchmarks and represent three distinct modeling paradigms.
These three models, as employed in analytical studies such as ~\cite{analysis/klenitskiy2023turning, analysis/betello2024investigating}, generate predictions from a single sequence-level representation.
This shared prediction interface allows us to directly compare their prediction-time behaviors across different inductive biases.
In addition to these standard baselines, we incorporate two models to extend our analysis:

\begin{itemize}[leftmargin=1.1em]

    \item \textbf{SASRec}~\cite{model/kang2018self} represents causal self-attention–based Transformers, in which next-item prediction relies on the final-position representation under a strictly autoregressive information flow.
    
    \item \textbf{GRU4Rec}~\cite{model/rnn/hidasi2015session} represents recurrent architectures, where user preferences are incrementally aggregated through sequential hidden-state transitions.

    \item \textbf{BERT4Rec}~\cite{model/sun2019bert4rec} represents bidirectional self-attention–based Transformers, which leverage both past and future context through an item-masking objective.
    At inference time, we predict by masking the target position.

     \item \textbf{DuoRec}~\cite{model/duorec/qiu2022contrastive} is a representative model that builds upon SASRec by incorporating a contrastive learning loss. We utilize this model to analyze how modifications to the loss function affect the model's dependency on the final item.

    \item \textbf{BSARec}~\cite{model/BSARec/shin2024attentive} is a recently proposed model that, while utilizing a self-attention mechanism similar to SASRec, operates within the frequency domain. 
\end{itemize}

Implementation details of each model and the training environment are provided in our repository.

\subsubsection{Evaluation Metrics}
\label{subsec:metrics}
Following ~\cite{analysis/gusak2025time}, we evaluate all models under a Global Time Split (GTS) protocol and report NDCG@K and HR@K on the test set; we set $K=10$.
Prior work has shown that leave-one-out evaluation allows future interactions to be included in the training set, resulting in temporal information leakage and overly optimistic performance estimates~\cite{analysis/GTS/sun2023take,analysis/hidasi2023widespread,analysis/gusak2025time}.
To avoid this issue, we adopt a GTS protocol, which strictly separates training and testing interactions in time.
Specifically, we employ the \emph{GTS-Last} target strategy together with a GT-based validation split, where the training and validation sets jointly account for 90\% of the data and the remaining portion is reserved for testing.

Finally, we report results using \emph{full-catalog ranking}, in which each model ranks the ground-truth item against the entire item set.
This choice avoids sampling-based metrics, which have been shown to yield inconsistent results and potentially misleading model comparisons~\cite{analysis/metric_sample/dallmann2021case,analysis/metric_sample/krichene2020sampled,analysis/hidasi2023widespread}.
To ensure a fair evaluation of next-item prediction, we exclude previously interacted items from the top-K recommendation lists (i.e., filter-seen)~\cite{analysis/filterseen/higley2022building}. However, we bypass this filtering process for Diginetica and Zvuk, as repeat consumption is a characteristic feature of user behavior in these specific datasets~\cite{analysis/klenitskiy2024does}.

\section{Empirical Characterization of Last-Item Reliance}
\label{sec:empirical}

In this section, we empirically characterize the inference-time behavior of SASRec~\cite{model/kang2018self}, with a particular focus on its tendency to rely on the last item in the input sequence.
Using a diverse set of datasets, we characterize the structure of inference-time behavior through two complementary diagnostic tools: inference-time position shuffling and a recency-localized hit-rate metric.

\begin{figure*}[t]
  \centering
  \includegraphics[width=\linewidth]{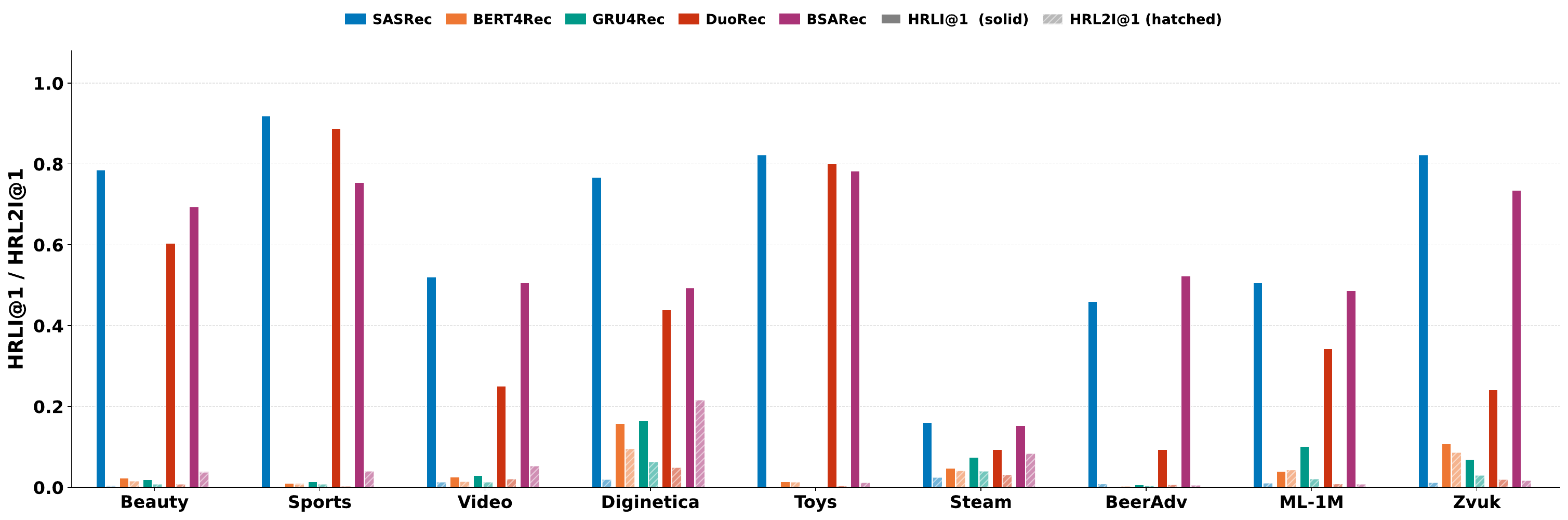}
  \caption{HRLI@1 and HRL2I@1 across models and datasets (computed without filter-seen). HRLI@1 measures whether the last interacted item $v_t$ appears in the top-1 list; HRL2I@1 is defined analogously for $v_{t-1}$.}
  \Description{
A grouped bar chart compares HRLI at 1 and HRL2I at 1 for five models across nine datasets: Beauty, Sports, Video, Diginetica, Toys, Steam, BeerAdvocate, ML-1M, and Zvuk. For each model and dataset, a solid bar represents the frequency with which the last interacted item is ranked first, and a hatched bar represents the corresponding frequency for the second-to-last item. SASRec, DuoRec, and BSARec generally show tall solid bars and very small hatched bars, producing a large gap between HRLI and HRL2I. The largest solid bars occur for several sparse datasets, including Beauty, Sports, and Toys, as well as Zvuk. BERT4Rec and GRU4Rec generally show lower HRLI values and smaller differences between the two positional metrics. Diginetica and Steam exhibit comparatively smaller last-item localization than several other datasets, although the solid bars for the causal self-attention models remain higher than their hatched counterparts.
}
  \label{fig:hrli}
\end{figure*}


\subsection{Inference-Time Position Shuffling}
\label{sec:shuffle}

We first verify position sensitivity via controlled input perturbations, which serve as a counterfactual probe of positional influence.
Specifically, given an interaction sequence, we construct variants where (i) only the final item is preserved while earlier positions are randomly permuted, and (ii) the entire sequence is permuted.
Comparing prediction performance across these variants isolates the effect of the final position.

\paragraph{Observations.}

Figure~\ref{fig:shuffle} illustrates representative results for
sparse, dense, and weakly sequential datasets. Across the complete
nine-dataset results, SASRec exhibits three distinct behavioral
regimes characterized by data density and sequentiality.

On sparse datasets, SASRec is largely insensitive to partial shuffling but shows a sharp performance drop under full shuffling. This indicates that the final item contains useful signals and that SASRec's predictions are heavily dependent on it. 
On denser datasets such as ML-1M, while partial shuffling leads to a moderate decrease in performance, a much more substantial drop is still observed under full shuffling.

For datasets with inherently weak sequential signals, such as Steam and Diginetica~\cite{analysis/klenitskiy2024does}, shuffling has a negligible effect on all models. 
These observations align with prior findings~\cite{analysis/xu2026markovian}; full results for all datasets and additional model variants, including DuoRec and BSARec, are available in our repository. DuoRec and BSARec also exhibit trends similar to SASRec.

\subsection{Quantifying Last-Item Reliance via HRLI}
\label{sec:hrli}

While inference-time shuffling probes a model’s sensitivity to perturbations in input order, it does not directly characterize how the predicted ranking is structured under standard inference.
To more directly quantify positional bias in the predicted ranking, we adopt the \emph{Hit Rate of the Last Item} (HRLI@K)~\cite{analysis/oh2024measuring} as a diagnostic measure.
This metric enables a systematic and quantitative characterization of positional bias, and will be used throughout the subsequent analysis.

HRLI@K is defined as
\begin{equation}
\mathrm{HRLI}@K
=
\frac{1}{|\mathcal{D}|}
\sum_{S_u \in \mathcal{D}}
\mathbb{I}\bigl(v_t \in \mathrm{Top}\text{-}K(S_u)\bigr),
\end{equation}
where $\mathcal{D}$ denotes the evaluation set of user sequences,
$\mathrm{Top}\text{-}K(S_u)$ is the recommendation list induced by the predicted scores
$p(v_{t+1}=v \mid S_u; \Theta)$,
and $\mathbb{I}(\cdot)$ is the indicator function that equals $1$ if the condition is satisfied and $0$ otherwise.
Intuitively, HRLI@K measures the fraction of user sequences for which the last item in the input sequence reappears in the Top-$K$ recommendation list.

Similarly, we define the \emph{Hit Rate of the Second-to-Last Item} (HRL2I\allowbreak@\allowbreak K) by replacing $v_t$ with the second-to-last item $v_{t-1}$.
Together, HRLI and HRL2I quantify how frequently items at specific positions in the input sequence are recovered in the predicted ranking.

In contrast to the standard ranking metrics used in Section~\ref{subsec:metrics}, HRLI and HRL2I are computed without applying filter-seen across all datasets.
This choice follows directly from the definition of these metrics, which are designed to measure how frequently a specific item appearing at a given position in the input sequence is ranked by the model.
Applying filter-seen would trivially prevent the target item from appearing in the recommendation list, rendering HRLI and HRL2I ill-defined or uninformative.
Although this evaluation setting differs from typical deployment scenarios, it is necessary to faithfully capture the intended quantity measured by these diagnostics.
HRLI should not be interpreted as a direct counterfactual measure
of dependence on the final item. Rather, it measures how frequently
the model ranks an item from a specified input position within the
recommendation list under standard inference. We therefore use HRLI
as a recency-localized ranking diagnostic that complements the
perturbation-based analysis in Section~\ref{sec:shuffle}. Its interpretation requires additional care in repeat-consumption domains.

Throughout our experiments, we focus on the case $K=1$.
Under this setting, HRLI@1 captures the most extreme form of last-item localization, namely the case in which the last item is ranked first.
We adopt this setting for clarity, as it provides a sharp and easily interpretable signal of positional dominance.
Results for larger values of $K$ exhibit similar qualitative trends and are reported in our repository.

\paragraph{Observations.}
Figure~\ref{fig:hrli} shows HRLI@1 and HRL2I@1 values for each model.
Our results across nine datasets with varying domains and properties reveal that causal self-attention models, including SASRec, DuoRec, and BSARec, generally yield substantially higher HRLI@1 than HRL2I@1. Conversely, BERT4Rec and GRU4Rec display a much smaller gap between these metrics, despite a slight inclination toward HRLI@1. These results reveal a highly localized ranking tendency toward
the last interacted item in SASRec-based models.
Even in datasets such as Steam and Diginetica, HRLI consistently exceeds HRL2I in relative terms. This provides a complementary perspective to our shuffling experiments, revealing a degree of last-item reliance that remained undetected by sensitivity-based diagnostics.

Taken together, these complementary diagnostics provide consistent
evidence of highly localized last-item behavior in SASRec, but remain agnostic to the mechanisms responsible for this behavior.
To move beyond descriptive evidence, we next analyze how information from different positions is structurally composed within the attention block.



\begin{figure*}[t]
  \centering
  \includegraphics[width=\linewidth]{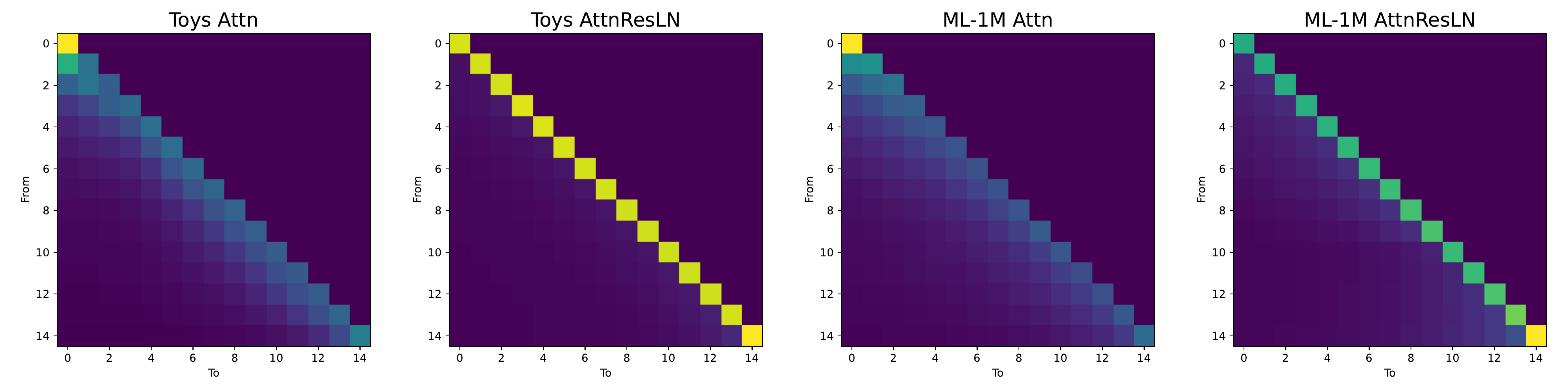}
\caption{
Input-wise contribution patterns in SASRec for the representative
sparse and dense datasets, Toys and ML-1M. From left to right:
Toys (Attn), Toys (AttnResLN), ML-1M (Attn), and ML-1M
(AttnResLN). Compared with the attention-only contributions
(Attn), the complete attention block outputs after residual addition
and layer normalization (AttnResLN) exhibit markedly stronger
diagonal components, indicating stronger preservation of
same-position information.
}
  \label{fig:attn_patterns_beauty}
  \Description{
Four heatmaps visualizing position-wise contribution patterns for SASRec on the Toys and ML-1M datasets.
From left to right, the figure shows attention-only outputs (Attn) and full attention-block representations including residual connections and layer normalization (AttnResLN) for Toys and ML-1M.
Compared to attention-only outputs, the full attention block exhibits stronger diagonal components, indicating increased preservation of self-information.
}

\end{figure*}

\section{Structural Analysis of Last-Item Reliance}
\label{sec:structural_analysis}

The empirical analyses in Section~\ref{sec:empirical} reveal that SASRec exhibits last-item reliance, as evidenced by both inference-time shuffling and HRLI diagnostics.
To examine how this behavior is structurally expressed in the
representation used for prediction, we analyze the contributions of
different input positions to the attention block output.

\subsection{Analysis of Attention Block}


A common approach to analyzing self-attention models is to inspect attention weights~\cite{analysis/wang2025your}.
However, attention weights alone are insufficient to explain the observed behavior.
In particular, attention weights typically vary smoothly across positions, whereas the last-item reliance observed in Section~\ref{sec:empirical} emerges as a highly discontinuous phenomenon.
This discrepancy suggests that ``where the model’s attention is placed'' does not necessarily reflect ``which information is actually used''.

This limitation arises because attention weights alone do not indicate which information is actually utilized in the final representation.
While attention weights describe how strongly each position is attended to, they do not directly quantify how much information from that position survives in the representation used for prediction.
In practice, information aggregation within a sequence is influenced not only by the attention mechanism, but also by other architectural components, most notably residual connections.
To address this issue, we analyze the full attention block, explicitly accounting for all its constituent components.

Specifically, we adopt a norm-based analysis~\cite{analysis/kobayashi2020attention,analysis/kobayashi2021incorporating}.
This approach quantifies each position's contribution using the norm of its input-wise component in the block output, rather than relying on attention weights alone. It thereby accounts for variations in the norms of transformed value vectors, residual addition, and layer normalization.



Formally, let $x_i$ denote the input representation at position $i$. The output of a causal self-attention block is:
\[
\tilde{x}_i = \text{LN}(\text{Attn}(x_i, X) + x_i).
\]

Following~\cite{analysis/kobayashi2020attention,analysis/kobayashi2021incorporating}, this representation can be decomposed into input-wise contributions:
\[
\tilde{x}_i = \sum_j F_i(x_j),
\]
where \(F_i(x_j)\) denotes the contribution of position \(j\) to position \(i\).

We quantify contextual aggregation using the mixing ratio:
\[
r_i = \frac{\sum_{j \ne i} \|F_i(x_j)\|}{\sum_j \|F_i(x_j)\|}.
\]

Lower values indicate dominance of self-information retention, while higher values suggest stronger aggregation of contextual information through the attention mechanism.
Mixing ratios are computed using norm-based decomposition of the full
attention block, in which multi-head attention is already integrated
into a single representation and thus requires no explicit head-wise
averaging. Table~\ref{tab:mixing_ratios_combined} reports values averaged across positions, layers, and test sequences to provide a compact cross-dataset comparison. To examine whether residual dominance is specific to the final position or concentrated in a particular layer, we additionally report position- and layer-specific AttnResLN mixing ratios for representative sparse and dense datasets in
Table~\ref{tab:position_layer_mixing}. Because representations at deeper layers incorporate transformations from preceding layers, these values characterize the representation after each layer rather than the isolated incremental effect of that layer.
Since this approach is applicable to attention-based models with
a compatible block structure regardless of their learning
objective, we extend our analysis to DuoRec as well.
Details of the derivation are provided in the supplementary material
in our repository.~\footnote{https://github.com/keito0329/Residual/blob/main/supplement.pdf}

\begin{table}[t]
\centering
\small 
\setlength{\tabcolsep}{4pt} 
\caption{Mixing ratios of SASRec and DuoRec.
Attn denotes the attention-only output, +Res denotes the output
after residual addition, and +LN denotes the subsequent output
after layer normalization.}
\label{tab:mixing_ratios_combined}
\begin{tabular}{l ccc ccc}
\hline
& \multicolumn{3}{c}{\textbf{SASRec}} & \multicolumn{3}{c}{\textbf{DuoRec}} \\
\cmidrule(lr){2-4} \cmidrule(lr){5-7}
\textbf{Dataset} & \textbf{Attn} & \textbf{+Res} & \textbf{+LN} & \textbf{Attn} & \textbf{+Res} & \textbf{+LN} \\ \hline
Beauty       & 0.776 & 0.119 & 0.119 & 0.776 & 0.116 & 0.115 \\
Sports       & 0.775 & 0.126 & 0.122 & 0.790 & 0.119 & 0.119 \\
Toys         & 0.772 & 0.123 & 0.123 & 0.780 & 0.121 & 0.121 \\
Video        & 0.798 & 0.169 & 0.169 & 0.773 & 0.170 & 0.170 \\
Diginetica   & 0.721 & 0.150 & 0.149 & 0.724 & 0.152 & 0.151 \\
ML-1M        & 0.856 & 0.395 & 0.386 & 0.862 & 0.367 & 0.359 \\
Steam        & 0.794 & 0.317 & 0.276 & 0.767 & 0.299 & 0.277 \\
BeerAdvocate & 0.959 & 0.301 & 0.287 & 0.949 & 0.311 & 0.307 \\
Zvuk         & 0.941 & 0.267 & 0.265 & 0.926 & 0.274 & 0.271 \\
\hline
\end{tabular}
\end{table}

\subsection{Results and Structural Interpretation}

Table~\ref{tab:mixing_ratios_combined} reports the mixing ratios computed at different stages of the attention block: the attention output alone (\textbf{Attn}), after adding the residual connection (\textbf{AttnRes}), and after applying layer normalization (\textbf{AttnResLN}).

Across all datasets, the Attn values are consistently high, indicating that the attention output aggregates substantial information from other positions. 
Once the residual connection is added, however, the mixing ratio drops sharply. This pattern reveals a consistent structural
tendency: although contextual information is aggregated by the
attention operation, the residual addition sharply shifts the balance of the block output toward the same-position contribution.

The additional effect of layer normalization is relatively minor
compared with the effect of residual addition.
Across datasets, the AttnRes and AttnResLN values are nearly identical, suggesting that residual addition is the primary structural factor associated with the reduction in contextual mixing.
While the absolute values of the mixing ratio vary across datasets, the qualitative trend is consistent.

To illustrate this tendency, Figure~\ref{fig:attn_patterns_beauty} visualizes the input-wise contribution patterns in SASRec for the Toys and ML-1M datasets, comparing the attention output alone with the complete attention block output after residual addition and layer normalization.
Compared with the attention-only contribution patterns, the full block exhibits markedly stronger diagonal components, indicating strong preservation of same-position information.


\begin{table}[t]
    \centering
    \caption{Position- and layer-specific AttnResLN mixing ratios of
SASRec on Toys and ML-1M, representing sparse and dense settings,
respectively. Values are averaged over test sequences.}
    \label{tab:position_layer_mixing}
    \small
    \begin{tabular}{lcccc}
        \toprule
        Dataset & Layer & $L$ & $L-1$ & $L-2$ \\
        \midrule
        Toys  & 1 & 0.1278 & 0.1283 & 0.1251 \\
              & 2 & 0.1182 & 0.1178 & 0.1135 \\
        \midrule
        ML-1M & 1 & 0.3984 & 0.3819 & 0.3798 \\
              & 2 & 0.3660 & 0.3617 & 0.3606 \\
        \bottomrule
    \end{tabular}
\end{table}

Table~\ref{tab:position_layer_mixing} further examines whether residual dominance is specific to the final position or concentrated in a particular layer. 
Within each layer, the mixing ratios vary only modestly across $L$, $L-1$, and $L-2$. 
The maximum absolute position-wise difference is 0.0032 in layer 1 and 0.0047 in layer 2 on Toys, and 0.0186 and 0.0054, respectively, on ML-1M.
Notably, the final-position mixing ratio is not consistently lower
than those of the preceding positions; on ML-1M, it is slightly higher.
Together with the sharp reduction in mixing observed after residual addition in Table~\ref{tab:mixing_ratios_combined}, these results indicate that residual dominance is broadly expressed across the examined positions and in both Transformer layers, rather than being specifically concentrated at the final position.

These results refine the structural account of last-item reliance.
The attention block exhibits a broadly self-preserving tendency across the examined positions. 
In SASRec, however, next-item prediction is formed exclusively from the final-position representation. 
The preserved same-position component at this prediction interface is
therefore anchored to the most recently interacted item. 
Consequently, the discontinuous last-position privilege observed in Section~\ref{sec:empirical} can be understood as the interaction between this broadly self-preserving tendency and the final-position-only prediction interface.


\subsection{Discussion}

In principle, the norm-based analysis employed in this section can be applied to other Transformer-based sequential recommendation models, including bidirectional models such as BERT4Rec.
However, we deliberately refrain from conducting a structural comparison with BERT4Rec in this work.
In SASRec-style causal models, predictions are made from the encoder output at the final item position, whose residual input is associated
with an actual interacted item.
Consequently, the mixing ratio reflects a structural trade-off between preserving the final-position input, which is anchored to the last interacted item, and aggregating information from preceding positions.
In contrast, BERT4Rec predicts the next item from the encoder output at a \texttt{[MASK]} position, whose residual input corresponds to a virtual placeholder rather than an observed item.
As a result, the decomposition into preserving versus mixing effects does not align directly with last-item reliance in sequential recommendation, making the mixing ratio less task-aligned for structural interpretation.
For this reason, we restrict our structural analysis to SASRec-style causal models, where the interpretation of information preservation and aggregation is clearer and more directly connected to the prediction process.

\begin{figure*}[t]
  \centering
  \includegraphics[width=\linewidth]{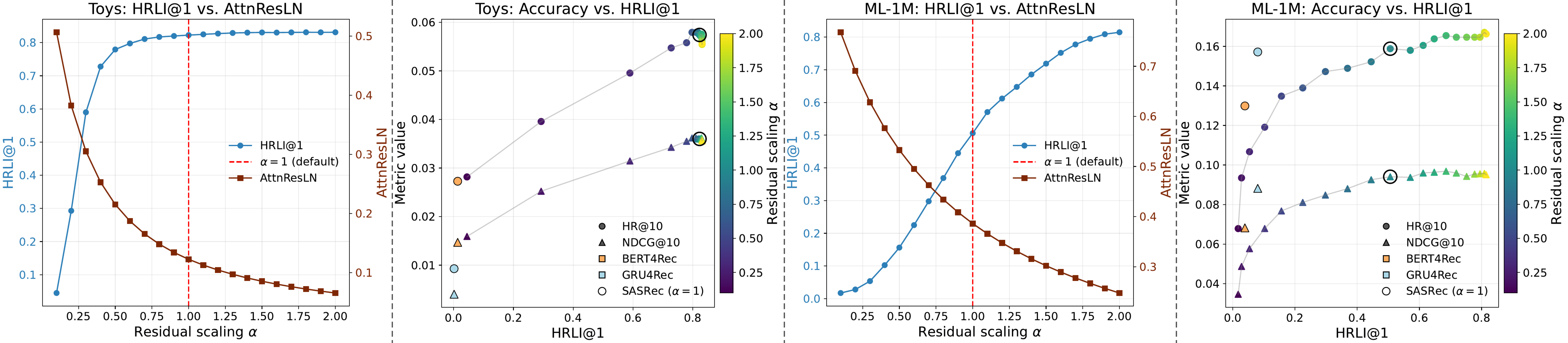}
  \caption{Effect of inference-time residual scaling ($\alpha$) applied to the residual branch at all sequence positions in SASRec.
Left: structural mixing averaged over positions and layers after the full attention block (AttnResLN; mixing ratio) and recency-localized reliance (HRLI@1) as functions of $\alpha$; the red dashed line indicates the standard setting $\alpha=1$.
Right: accuracy--reliance trade-off, plotting HR@10 and NDCG@10 against HRLI@1; each colored point corresponds to a value of $\alpha$ (colorbar), and baseline markers indicate BERT4Rec and GRU4Rec.
}
  \label{fig:tradeoff}
  \Description{Effect of inference-time residual scaling on structural mixing, recency bias, and accuracy.}
\end{figure*}

\section{Inference-Time Probing of Residual Contribution}
\label{sec:Inferencetime}

While Section~\ref{sec:structural_analysis} provides a structural account of last-item reliance,
it remains unclear whether prediction-time representations and
rankings vary systematically with residual strength.
We therefore apply inference-time residual scaling as a controlled
mechanistic probe and examine its relationship to contextual mixing,
last-item reliance, and ranking outcomes.
This intervention is intended as a sensitivity analysis rather than
strict causal identification.

\subsection{Residual Scaling as a Mechanistic Probe}
\label{subsec:control_design}

In a SASRec-style Transformer block, the representation after self-attention and residual addition is
\begin{equation}
h_i = x_i + \mathrm{Attn}(x)_i.
\end{equation}

We modify this computation at inference time:
\begin{equation}
h_i^{(\alpha)} = \alpha x_i + \mathrm{Attn}(x)_i,
\end{equation}
where $\alpha$ modulates the strength of residual preservation. Reducing $\alpha$ weakens same-position residual preservation at
all sequence positions in every Transformer layer and increases the relative contribution of contextual aggregation. 
At the final-position prediction interface, this reduces the direct preservation of information anchored to the last item.

As shown in Figure~\ref{fig:tradeoff}, decreasing $\alpha$
consistently increases the AttnResLN mixing ratio and decreases HRLI@1.
Residual strength is therefore monotonically associated with both
contextual mixing and last-item reliance, supporting the structural
account in Section~\ref{sec:structural_analysis}.

We further analyze how ranking performance changes with $\alpha$. As $\alpha$ decreases, HR@10 and NDCG@10 consistently degrade, while last-item reliance is reduced. This reveals a trade-off between preserving self-information at the final position and incorporating contextual information from preceding positions.

Importantly, this trade-off does not by itself indicate whether contextual information contributes meaningful predictive signals or merely reflects degradation of the final representation. These possibilities cannot be distinguished using standard ranking metrics alone. Therefore, at this stage, residual scaling establishes that prediction behavior varies systematically with residual strength, but does not determine whether additional
predictive signals are effectively utilized. To address this question, we next examine whether useful predictive signals exist beyond the final position and whether reducing residual strength can recover some of these
final-position misses.
Full residual-scaling curves for all datasets are available in our repository.

\begin{table*}[t]
\centering
\caption{Correct non-final predictions among final-position misses.
The first row reports the standard HR@10 obtained from the
final-position representation. For each preceding position $L-k$,
the remaining rows report the joint proportion of all test sequences
for which the final-position prediction misses the ground-truth item,
while the representation at position $L-k$ ranks it within the top 10.
These values indicate the prevalence of potentially useful predictive
signals at non-final positions.}
\label{tab:recovery_preceding_positions}
\setlength{\tabcolsep}{4pt} 
\footnotesize
\begin{tabular}{lccccccccc}
\toprule
 & \multicolumn{9}{c}{Dataset} \\
\cmidrule(lr){2-10}
Metric & Beauty & Sports & Toys & Video & Diginetica & ML-1M & Steam & BeerAdvocate & Zvuk \\
\midrule
HR@10 at $L$ (standard inference)
& 0.0381 & 0.0342 & 0.0574 & 0.0403 & 0.2900 & 0.1588 & 0.0639 & 0.0389 & 0.2416 \\
HR@10 at $L-1$ and \text{miss at } $L$ 
& 0.0116 & 0.0157 & 0.0185 & 0.0171 & 0.1153 & 0.0347 & 0.0235 & 0.0150 & 0.0583 \\
HR@10 at $L-2$ and \text{miss at } $L$ 
& 0.0124 & 0.0125 & 0.0134 & 0.0145 & 0.0782 & 0.0422 & 0.0241 & 0.0141 & 0.0415 \\
HR@10 at $L-3$ and \text{miss at } $L$ 
& 0.0112 & 0.0115 & 0.0150 & 0.0124 & 0.0914 & 0.0298 & 0.0236 & 0.0139 & 0.0431 \\
\bottomrule
\end{tabular}
\\[0.5ex]
\end{table*}

\subsection{Correct Predictions at Non-Final Positions}
\label{sec:recovery_preceding}

Here, a non-final hit means that the representation at position
$L-k$ ranks the final test target $v_{t+1}$ within the top 10; it
does not refer to the position-specific next-item target used
during training.

Table~\ref{tab:recovery_preceding_positions} reports the joint
proportion of all test sequences for which the final-position
prediction misses the ground-truth item, while the representation at a preceding position ranks it within the top 10. The results show that correct predictions
can occur at non-final positions even when the final-position
prediction fails.

Across datasets, the largest single-position joint rate corresponds
to approximately 24--46\% of the standard HR@10 value, indicating
that such cases occur at a non-negligible scale. These results
establish the presence of potentially useful predictions at non-final
positions. In the next subsection, we examine whether reducing
residual strength can recover the corresponding final-position
misses.

\subsection{Recovery of Final-Position Misses}
\label{subsec:recovery_scaling}





To test whether reducing residual strength can recover
final-position misses with correct non-final hits, we consider
sequences satisfying the following conditions under the standard
setting $\alpha=1$: (i) the final-position representation fails to
rank the final test target $v_{t+1}$ within the top 10, and
(ii) at least one representation at $L-1$, $L-2$, or $L-3$ ranks
$v_{t+1}$ within the top 10. This conditioning isolates
final-position misses for which a correct non-final hit is already
present under standard inference.

\begin{figure}[t]
  \centering
  \includegraphics[width=0.65\linewidth]{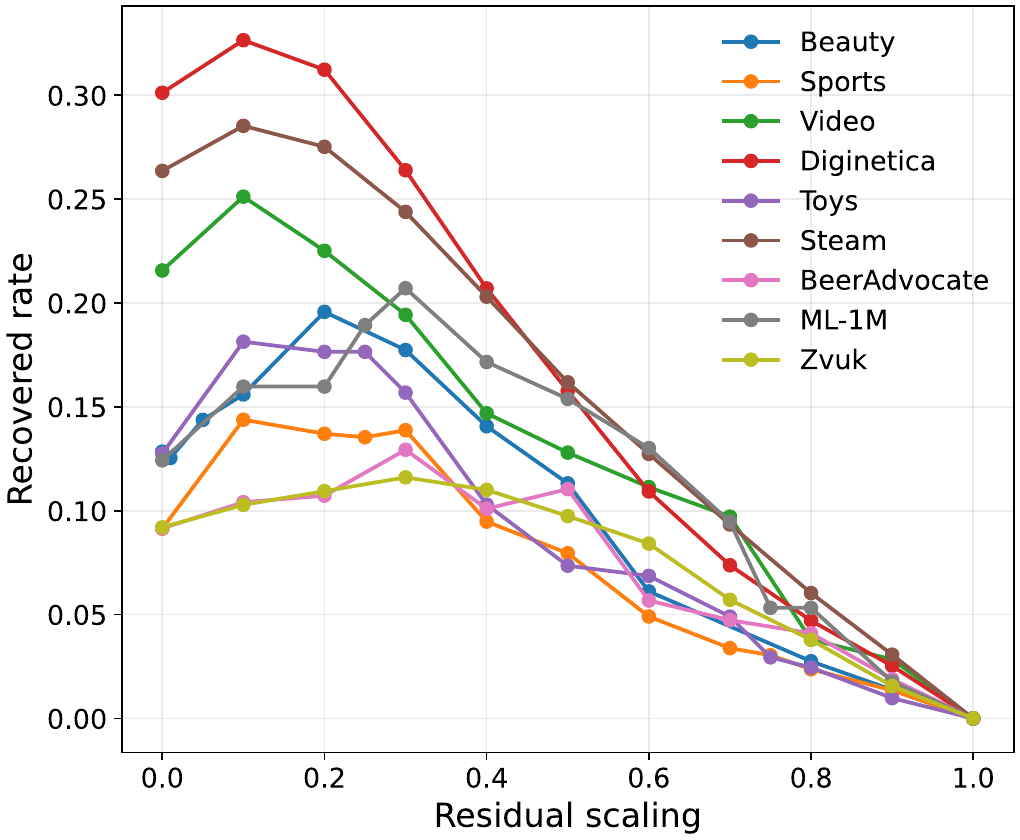}
\caption{Recovery of final-position misses under inference-time
residual scaling. For each dataset, we consider sequences for
which the standard final-position inference ($\alpha=1$) fails to
rank the ground-truth item within the top 10, while at least one
representation at $L-1$, $L-2$, or $L-3$ ranks it within the
top 10. The curve reports the fraction of such sequences for which
the final-position representation ranks the ground-truth item
within the top 10 after applying residual scaling with factor
$\alpha$ at all sequence positions. The recovery rate is zero at
$\alpha=1$ by definition.}
\Description{
A multi-line plot shows the recovery rate of final-position misses as a function of the residual scaling factor alpha for nine datasets. The horizontal axis ranges from complete removal of the residual contribution at alpha equal to zero to the standard setting at alpha equal to one. The vertical axis shows the fraction of conditioned sequences that are recovered at the final position. All curves reach zero at alpha equal to one by construction. For most datasets, recovery increases as alpha is reduced from one, reaches its maximum at an intermediate value around 0.1 to 0.3, and then remains stable or decreases slightly as alpha approaches zero. Diginetica shows the highest peak recovery, exceeding 0.3, while the remaining datasets exhibit lower but consistently positive peaks. The common non-monotonic pattern indicates that moderate residual reduction generally recovers more cases than either the standard residual strength or complete residual removal.
}
\label{fig:recover}
\vspace{-0.3cm}
\end{figure}

For each value of $\alpha$, we measure the fraction of these
sequences for which the final-position representation ranks
$v_{t+1}$ within the top 10. As shown in
Figure~\ref{fig:recover}, decreasing $\alpha$ recovers a subset of
final-position misses, with peak recovery observed around
$\alpha \approx 0.1$--$0.3$.

Because every evaluated sequence contains a correct non-final hit
under standard inference, the observed recovery is consistent with
the hypothesis that reducing residual preservation increases the
contribution of non-final predictive information to the
final-position ranking.

\subsection{Validation-Based Selection}
\label{subsec:validation_selection}

While our primary goal is analysis rather than optimization, we include this experiment to demonstrate that the observed controllability is reproducible without access to test data. We adopt a simple validation-based selection rule:

\begin{equation}
\begin{aligned}
\alpha^* = \arg\max_{\alpha} \quad & \mathrm{Recovery}_{\text{val}}(\alpha) \\
\text{s.t. } \quad & \mathrm{HR@10}_{\text{val}}(\alpha) \ge (1-\varepsilon)\mathrm{HR@10}_{\text{val}}(1)
\end{aligned}
\end{equation}

\begin{figure}[t]
  \centering
  \includegraphics[width=1.0\linewidth]{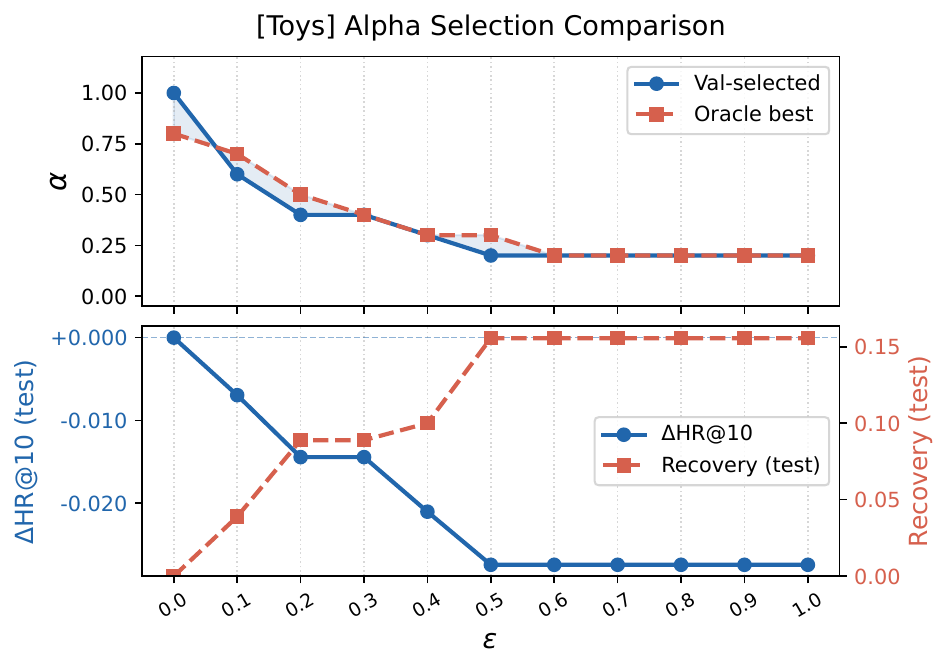}
  \caption{Validation-based reproduction of the accuracy–recovery trade-off. (Top) Comparison between validation-selected and oracle-selected $\alpha$ across different values of $\varepsilon$.
(Bottom) Corresponding test performance, showing the trade-off
between $\Delta$HR@10 relative to the $\alpha=1$ baseline and
recovery under validation-based selection.
On Toys, validation-based selection reproduces the same accuracy--recovery trade-off observed under oracle tuning.
}
\Description{
Two vertically stacked plots summarize validation-based residual-scaling selection on Toys. The top plot shows the selected residual scaling factor alpha as a function of the allowed accuracy-loss parameter epsilon. The validation-selected curve and the oracle-selected curve follow similar downward trajectories: selected alpha is high under a strict accuracy constraint, decreases as epsilon increases, and stabilizes near a small value under looser constraints. The bottom plot shows the resulting test-set trade-off. The left vertical axis reports the change in HR at 10 relative to standard inference, and the right vertical axis reports recovery. As epsilon increases, the HR at 10 change becomes more negative, while recovery increases. Both quantities stabilize once the selected alpha becomes approximately constant. The two panels together show that validation-based selection reproduces a trade-off similar to oracle selection without selecting alpha directly on the test set.
}
\label{fig:compare}
\vspace{-0.3cm}
\end{figure}

This formulation prioritizes recovery while constraining accuracy
degradation.
Figure~\ref{fig:compare} shows, on Toys, that validation-selected $\alpha$ closely matches oracle-selected $\alpha$ over a wide range of $\epsilon$ values, reproducing the same accuracy--recovery trade-off observed on the test set.

These results indicate that the controllability revealed by residual scaling is stable and can be reproduced using standard validation procedures.
However, this experiment is intended only as a demonstration of reproducibility and does not constitute a separate modeling contribution.
Results for additional datasets are available in our repository.

\textit{Summary.}
Taken together, the results in this section establish a consistent
picture: (1) prediction behavior varies systematically with residual
strength, (2) correct predictions can occur at non-final positions
when the final position misses, and (3) reducing residual strength
recovers a subset of these cases.

\section{Related Work}

\subsection{Sequential Recommendation}
Sequential recommender systems were initially dominated by RNN-~\cite{model/rnn/hidasi2015session,model/NARM/li2017neural} and CNN-based models~\cite{model/caser/tang2018personalized,model/CNN/yuan2019simple}, which improved upon Markov chain–based models~\cite{model/MC/zimdars2013using,model/MC/rendle2010factorizing} by capturing longer-term dependencies.
Transformer-based architectures later advanced the field by enabling flexible sequence modeling via self-attention~\cite{model/sun2019bert4rec,model/kang2018self,model/wu2020sse}, with causal self-attention models such as SASRec becoming widely adopted baselines.
Despite recent developments including graph-based methods~\cite{model/graph/qiu2020gag,model/graph/xu2019graph,model/graph/qiu2021exploiting}, generative models~\cite{model/diffusion/li2023diffurec,model/diffusion/yang2023generate,model/diffusion/mao2025distinguished}, and LLM-based approaches~\cite{model/llm/liu2024llm,model/llm/sakurai2025llm,model/llm/harte2023leveraging}, causal self-attention remains a core architectural choice in many sequential recommenders.
Understanding the inference behavior of such foundational models is therefore important for clarifying how sequential information is actually utilized.

\subsection{Attention Analysis}
Prior work has investigated which dependencies self-attention should emphasize in sequential recommendation~\cite{model/BSARec/shin2024attentive,analysis/wang2025your}.
Several studies~\cite{analysis/wang2025your,model/he2021locker} highlight the importance of local and short-term dependencies, showing that unconstrained global attention can be noisy or suboptimal, especially under sparse interaction regimes.
While these works analyze or modify attention mechanisms to improve performance, they largely focus on attention weights or attention modules in isolation.
In contrast, how attention outputs are utilized at prediction time remains underexplored, motivating our analysis of the full attention block with residual connections.

\subsection{Recency Bias}
Recency bias refers to the tendency of sequential recommender systems to rely disproportionately on recent interactions~\cite{analysis/chang2022recency}.
Prior work proposed both regularization-based mitigation strategies~\cite{analysis/chang2022recency} and evaluation metrics such as HRLI~\cite{analysis/oh2024measuring}, revealing that causal self-attention models often exhibit extreme last-item reliance.
More recent studies~\cite{analysis/xu2026markovian} further demonstrated that this behavior persists across architectures, embedding sizes, and loss functions, and argued that strong last-item reliance is essential for achieving high next-item prediction accuracy.

In contrast to these studies, which primarily characterize or justify last-item reliance as an empirical or task-level property, our work investigates how such extreme dependence is structurally
expressed in the prediction representation of SASRec.

\section{Conclusion}

We showed that causal self-attention recommenders often exhibit
highly localized last-item reliance at inference time.
Through norm-based analysis of the full attention block, we
identified residual dominance as a broadly self-preserving tendency
that, together with the final-position-only prediction interface,
provides a structural account of this behavior.
Inference-time residual scaling further revealed systematic
relationships among residual strength, contextual mixing, and
last-item reliance, and recovered a subset of final-position misses
with correct non-final hits.
These findings establish prediction-time information utilization as
a useful analytical lens for understanding causal self-attention
recommenders beyond attention weights and ranking accuracy alone.

\noindent\textbf{Limitations and future work.}
Residual dominance should not be interpreted as the sole cause of last-item reliance. 
Other factors, such as causal masking, positional information, the prediction interface, training objectives, data characteristics, and optimization may also contribute. 
Our analysis and intervention are restricted to causal self-attention models that predict from the final-position representation, and extending this characterization to other prediction heads and training objectives remains an important direction for future work. In addition, HRLI and HRL2I are designed as diagnostic metrics rather than replacements for standard accuracy metrics, and establishing their connection to online utility is an open problem. Finally, understanding how these findings translate to real-world recommendation settings remains an important direction for future work.

\begin{acks}
This work was partly supported by JSPS KAKENHI Grant Numbers
JP24K02942 and JP23K11141. 
\end{acks}

\bibliographystyle{ACM-Reference-Format}
\bibliography{sample-base}


\end{document}